\documentclass[10pt,letterpaper]{article}

\usepackage{tikz}
\usepackage{tikz-qtree}
\usepackage{cogsci}
\usepackage[anythingbreaks]{breakurl}
\usepackage{pslatex}
\usepackage{apacite}
\usepackage{pbox}
\usepackage{tabularx}
\usepackage{linguex}
\usepackage{subfig}
\usepackage{floatrow}
\usepackage{booktabs}
\definecolor{lightgreen}{RGB}{176, 245, 191}
\definecolor{darkred}{cmyk}{0.15,1,1,0}
\definecolor{darkblue}{cmyk}{1,0.9,0.1,0}
\newcommand{\depmem}[1] {\textbf{#1}}

\newcommand{\attractor}[1] {\textit{#1}}
\usepackage[colorinlistoftodos,prependcaption,textsize=scriptsize]{todonotes}

\title{Distinct patterns of syntactic agreement errors in recurrent networks and humans}
 
\author{{\large \bf Tal Linzen (tal.linzen@jhu.edu)} \\
  Department of Cognitive Science, Johns Hopkins University\AND
  {\large \bf Brian Leonard (bleonar10@jhu.edu)} \\
  Department of Cognitive Science, Johns Hopkins University}

\begin{document}

\maketitle

\begin{abstract}

    Determining the correct form of a verb in context requires an understanding of the syntactic structure of the sentence. Recurrent neural networks have been shown to perform this task with an error rate comparable to humans, despite the fact that they are not designed with explicit syntactic representations. To examine the extent to which the syntactic representations of these networks are similar to those used by humans when processing sentences, we compare the detailed pattern of errors that RNNs and humans make on this task. Despite significant similarities (attraction errors, asymmetry between singular and plural subjects), the error patterns differed in important ways. In particular, in complex sentences with relative clauses error rates increased in RNNs but decreased in humans. Furthermore, RNNs showed a cumulative effect of attractors but humans did not. We conclude that at least in some respects the syntactic representations acquired by RNNs are fundamentally different from those used by humans.

\end{abstract}

\section{Introduction}

Syntactic dependencies between words are most naturally expressed in terms of a structural representation of the sentence \cite{everaert2015structures}. The form of an English verb, for example, often depends on whether its subject is singular or plural. It is straightforward to identify the subject of a particular verb given a structural description of the sentence, but far from clear how to do so based only on the sequence of words. In particular, the subject is not necessarily the most recent noun preceding the verb; to determine the form of the verb in sentences such as \textit{the \textbf{key} to the cabinets \textbf{is} on the table}, it is necessary to ignore the linear proximity of \textit{cabinets} and reach back to the structurally relevant \textit{key}:

\bigskip

    \begin{tikzpicture}
        \centering
        \tikzset{level distance=20pt}
        {\small
        \Tree [.S [.NP [.D The ] [.NN \node{\depmem{key}}; ] [.PP to [.NP the [.NNS \node{\attractor{cabinets}}; ] ] ] ]  [.VP [.VB \node{\depmem{is}}; ] [.PP \edge[roof]; {on the table}  ] ] ]
        }
    \end{tikzpicture}

\bigskip

Models of human syntactic knowledge often include such explicit structural representations; aspects of these representations are often argued to be innate. By contrast, recurrent neural networks (RNNs) are not constructed with such pre-existing structural representations. These models must treat language learning as any other sequence learning problem; appropriate syntactic representations may \textit{emerge} in these networks through mere exposure to sentences. Proof-of-concept work examining whether such representations do in fact emerge was initially conducted on small synthetic languages \cite{elman1991distributed}. Recent technological advances have made it possible to revisit this question using larger networks trained on realistic corpora. \citeA{linzen2016assessing} trained RNNs to predict whether an upcoming verb should be singular or plural based on the words leading up to the verb (the \textbf{preamble}). The RNNs were trained on a large sample of sentences from Wikipedia, for example:\footnote{The only training signal was the number of the verb; the networks were not presented with the identity of the verb or with the continuation of the sentence after the verb, given here in brackets.}

\begin{figure}
    \centering
    \includegraphics[width=0.7\textwidth]{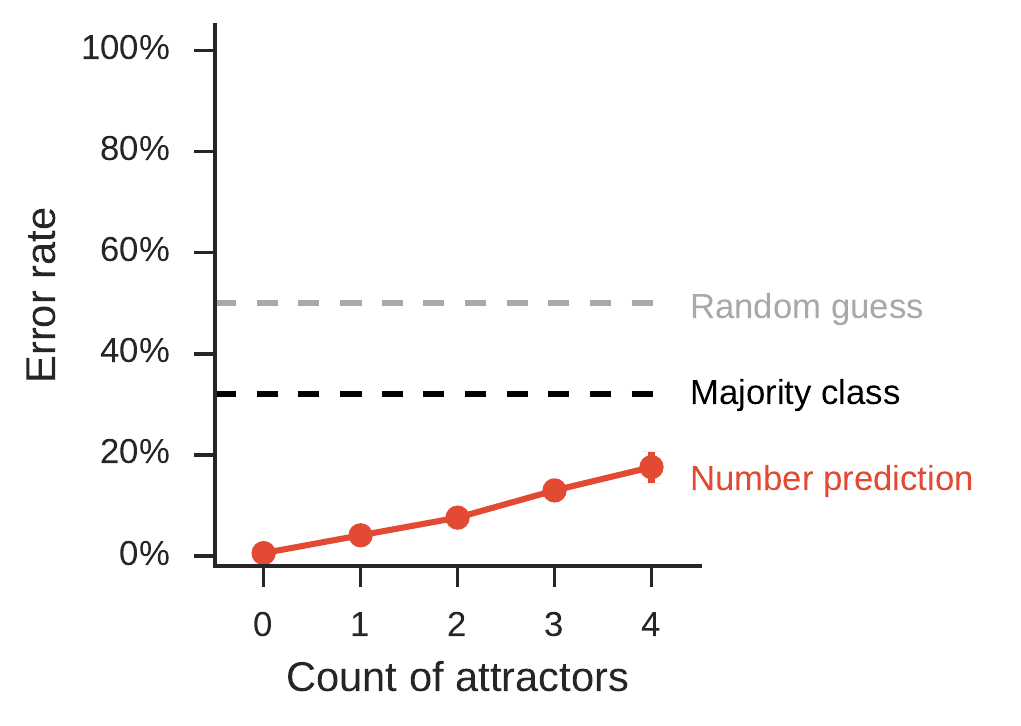}
    \caption{\label{fig:tacl}RNN error rates on verb number prediction as a function of the number of attractors, from \citeA{linzen2016assessing} (the majority class baseline always predicts a singular verb, with 68\% accuracy).}
\end{figure}

\ex.\label{ex:three_attractors}Yet the \depmem{ratio} of \attractor{men} who survive to the \attractor{women} and \attractor{children} who survive \depmem{is} [not clear in this story.]

\begin{table*}[!h]
    \centering
    \resizebox{0.8\textwidth}{!}{%
        \begin{tabular}{llll}
            \toprule
            Modifier type & Subject number & Local noun match & Preamble \\
            \midrule
            PP & Singular & Match & The demo tape from the popular rock singer \\
            PP & Singular & Mismatch & The demo tape from the popular rock singers \\
            PP & Plural & Match & The demo tapes from the popular rock singers \\
            PP & Plural & Mismatch & The demo tapes from the popular rock singer \\
            RC & Singular & Match & The demo tape that promoted the rock singer \\
            RC & Singular & Mismatch & The demo tape that promoted the rock singers \\
            RC & Plural & Match & The demo tapes that promoted the rock singers \\
            RC & Plural & Mismatch & The demo tapes that promoted the rock singer \\
            \bottomrule
        \end{tabular}%
    }
    \caption{\label{table:experiment1}Materials of Experiment~1 \cite{bock1992regulating}.}
\end{table*}

 The RNNs generalized this task well to new sentences: overall, they made an error in less than 1\% of the cases. Error rates increased in sentences with \textbf{attractors}, words of the opposite number from the subject that intervene between the subject and the verb, such as \textit{men} in~(1). Yet even in sentences with up to four attractors, RNNs performed better than simple baselines (Figure \ref{fig:tacl}). 

While the RNNs' degraded performance on sentences with attractors suggests that syntactic processing in RNNs is imperfect, it is not immediately clear that they differ from humans in this respect. It has long been known anecdotally that humans make occasional agreement errors. The first to investigate this phenomenon systematically were \citeA{bock1991broken}. Their participants listened to preambles (e.g., \textit{the key to the cabinets...}), and were asked to repeat those preambles and complete the sentence as they saw fit. This paradigm successfully elicited agreement errors; in particular, like the RNNs evaluated by \citeA{linzen2016assessing}, humans made many more agreement errors when the preamble contained an attractor. This raises the possibility that the syntactic representations that emerge in RNNs are similar to those used by humans to process language.

The goal of this paper is to carry out a detailed comparison between the agreement errors made by humans and RNNs. We focus on two factors: first, the type of syntactic structure that contains the attractor; and second, the structural position of the attractor. Our starting point is the comparison between the following sentence types \cite{bock1992regulating}:

\ex.\label{ex:prep}Prepositional phrase (PP):\\ The demo \textbf{tape} from the popular rock \textit{singers}...
    
\ex.\label{ex:rel}Relative clause (RC):\\ The demo \textbf{tape} that promoted the rock \textit{singers}...

Even though relative clauses are syntactically more complex than prepositional phrases, \citeA{bock1992regulating} found that attraction errors were somewhat \textit{less} common when the attractor was inside a relative clause, as in \ref{ex:rel}, than when it was in a prepositional phrase, as in \ref{ex:prep}. This finding has been taken to provide evidence for hierarchical representations in human sentence processing: the greater amount of abstract structure surrounding the relative clause insulates the subject from interference from nouns inside the clause. 

This paper is structured as follows. Experiment 1 is a conceptual replication of \citeA{bock1992regulating} using a paradigm comparable to the one on which RNNs can be easily tested. Comparing the results of Experiment 1 to a simulation using RNNs, we found significant qualitative similarities between the errors made by humans and RNNs; crucially, however, RNNs made many more errors in relative clauses than prepositional phrases, unlike humans. Experiment~2 found that RNN show a cumulative effect of multiple attractors whereas humans are only sensitive to the attractor nearest the subject. Using the materials of Experiment~2 and an additional set of materials, we demonstrate that the RNNs' increased difficulty with RCs is due to a faulty heuristic they learn, whereby RCs are assumed to be short.\footnote{The code and materials necessary to recreate the results in this paper will be made available at \url{https://github.com/jhupsycholing/RNNvsHumanSyntax}.}

\begin{figure*}
    \sidesubfloat[]{
        \includegraphics[width=0.45\textwidth]{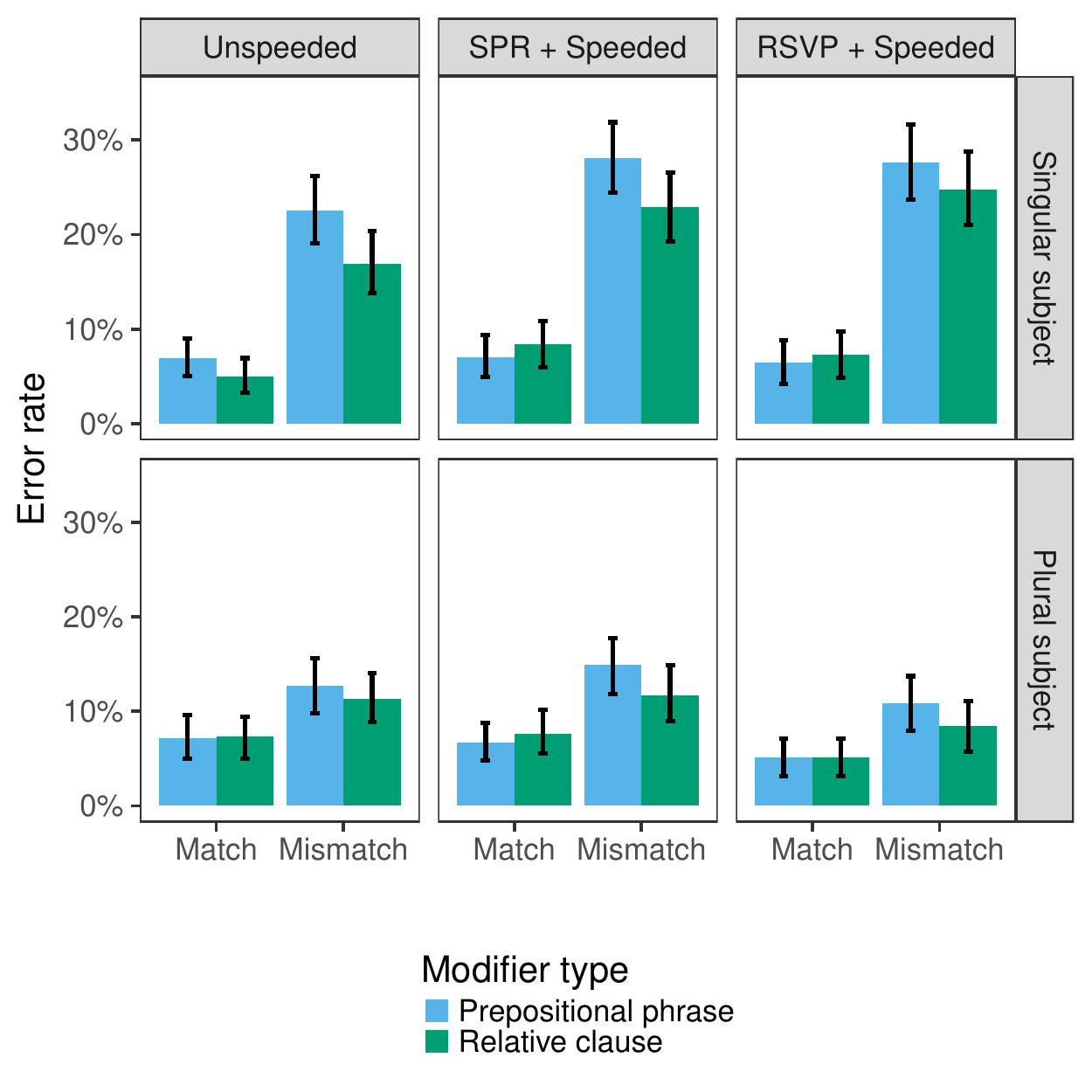}
        \label{fig:bc_rep}
    }
    \sidesubfloat[]{
        \includegraphics[width=0.45\textwidth]{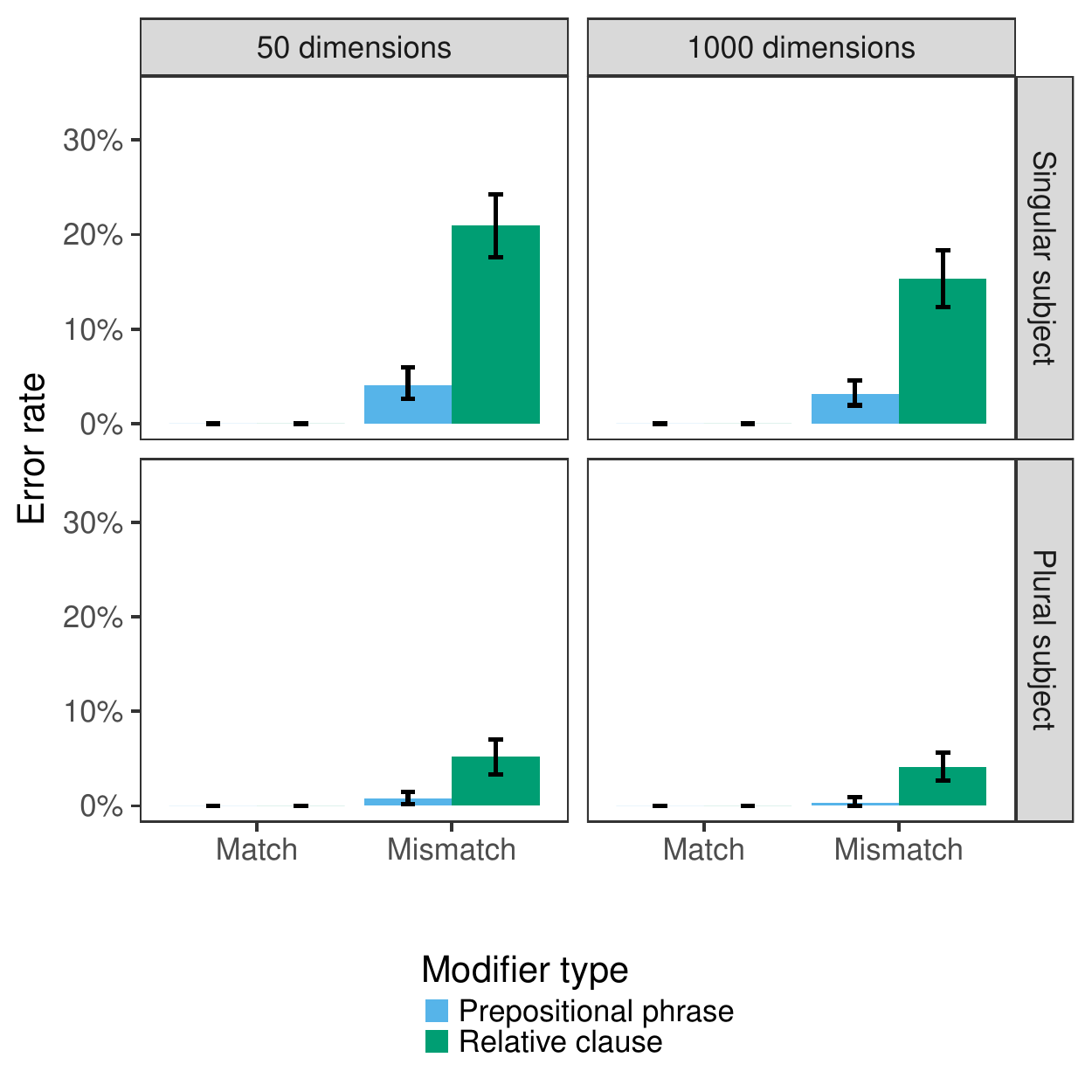}
        \label{fig:bc_rnn}
    }
    \caption{(a) Human agreement errors in Experiment~1. Error bars represent bootstrapped 95\% confidence intervals. (b) RNN predictions for the same materials. Error bars represent standard errors (across the 20 models in each category).}
\end{figure*}

\section{Experiment~1}

\citeA{bock1992regulating} had participants produce a sentence beginning with the preamble they heard. Following \citeA{staub2009interpretation}, we simplify this paradigm: our participants read the preamble and made a forced choice between a singular and a plural verb (specifically, \textit{is} and \textit{are}). We experiment with three variants of this methodology, designed to put participants under different degrees of time pressure. The hypothesis we had in mind was that under time pressure humans would be more likely to resort to suboptimal sequential strategies and produce errors similar to those produced by RNNs.

The three paradigms were as follows. In the \textbf{RSVP} paradigm, the words were displayed one by one in the center of the screen: each word was presented for 250 ms, followed by a blank screen displayed for 150 ms. The participants were then given 1500 ms to choose between the verbs \textit{is} and \textit{are} (presented in random order). In the \textbf{SPR} (self-paced reading) paradigm, the sentences were revealed word by word. Participants controlled the rate at which the words were revealed. As a word was revealed, previous words were replaced with strings of dashes. Participants assigned to this paradigm were again given 1500 ms to choose between \textit{is} and \textit{are} after the end of the sentence. Finally, in the \textbf{Untimed} paradigm, the full sentence and the two response options were revealed at the same time; participants were allowed to take as much as time they wished to make their choice.

Each participant in the experiment responded to 32 critical items and 56 filler items, for a total of 88 preambles. The critical items, in eight conditions, were drawn from \citeA{bock1992regulating} (see Table \ref{table:experiment1}). We created the filler items ourselves since those were not provided in the original paper. 

We recruited 384 participants (128 in each of the three paradigms) on Prolific.\footnote{\url{http://prolific.ac}} All participants were required to pass the platform's qualification for native English competence. We excluded participants who gave incorrect responses to more than 20\% of the fillers (Unspeeded: $n = 4$; SPR: $n = 4$; RSVP: $n = 15$).

The results are shown in Figure \ref{fig:bc_rep}. The qualitative pattern of results was very similar across the three paradigms, with a moderate reduction in error rates across the board in the Unspeeded paradigm. In all three experiments, participants made more errors when the local noun differed in number from the subject, the signature agreement attraction effect ($p < .001$; analyses performed using logistic mixed-effects models). We next examined the interactions with this attraction effect. Attraction errors were more common when the subject was singular ($p < .001$), although they were consistently produced with plural subjects as well ($p < .01$ within plural subjects in all three experiments). Attraction errors were significantly more frequent with PP than RC modifiers in the SPR experiment ($p = .02$) but not in the Unspeeded and RSVP experiments (Unspeeded: $p = .5$; RSVP: $p = .3$).

Finally, we combined the results of all three experiments, and focused in particular on sentences with a singular subject and a plural local noun, in which agreement attraction errors were most frequent. In this subset of the data, the effect of experiment was significant ($p = .02$), reflecting the somewhat lower proportion of errors in the Unspeeded paradigm. Errors were overall more likely after PPs than RCs ($p = .02$). The two factors did not interact ($p = .54$). Our data therefore does not provide evidence that greater time pressure makes the processing of complex sentences more error prone.

\paragraph{Human findings to be modeled:} Since there was no qualitative effect of time pressure, we will not attempt to model this factor in our simulations. Likewise, humans made agreement errors some fraction of the time (around 7\%) even in sentences without attractors (Match sentences); we will not attempt to capture these errors, which are likely due to lapses in attention. The qualitative patterns from the human experiments that we would like to compare to RNNs are:

\begin{enumerate}
    \item \textbf{Attraction:} Errors are more likely in the presence of an attractor.
    \item \textbf{Number asymmetry:} Errors are more likely when the subject is singular and the attractor is plural than the other way around.
    \item \textbf{Relative clause advantage:} Errors are somewhat more likely when the attractor is in a prepositional phrase than when it is in a relative clause.

\end{enumerate}

\section{Simulation of Experiment~1}

\paragraph{Model training:} We trained RNNs in a supervised way to predict the number of an upcoming verb, broadly following the protocol of \citeA{linzen2016assessing}.\footnote{RNNs perform somewhat worse when trained on word prediction (as ``language models''), without specific supervision on verb number prediction \cite{linzen2016assessing}, though in recent work they have been able to perform fairly well even in such a setting \cite{gulordava2018colorless}. We do not explore the word prediction training regimen here.} The corpus was extracted from the English Wikipedia, and the sentences were automatically parsed. We extracted sentences with present-tense third person verbs, the only context in which English verbs agree with their subjects. Dependencies in which the subject and the verb did not agree in number were excluded from consideration; such dependencies reflected either parse errors or agreement errors on the part of Wikipedia editors. We randomly selected one of the verbs in the sentence, and created a preamble by deleting all of the words from the verb on. The training corpus had about 1.27 million preambles in total. We set aside an additional 142000 preambles as a validation set.

\begin{figure*}
    \sidesubfloat[]{
        \includegraphics[width=0.45\textwidth]{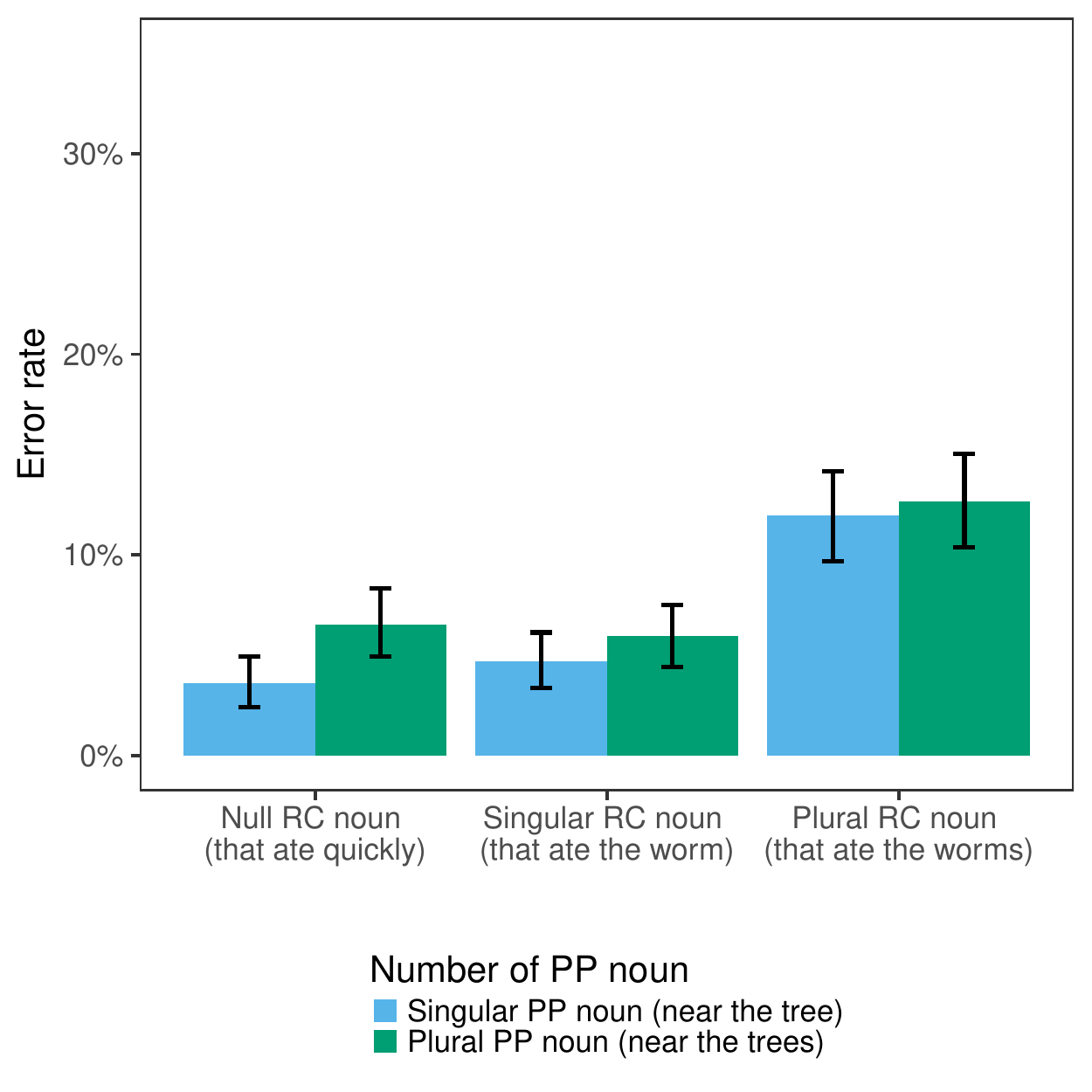}
        \label{fig:mult_rep}
    }
    \sidesubfloat[]{
        \includegraphics[width=0.45\textwidth]{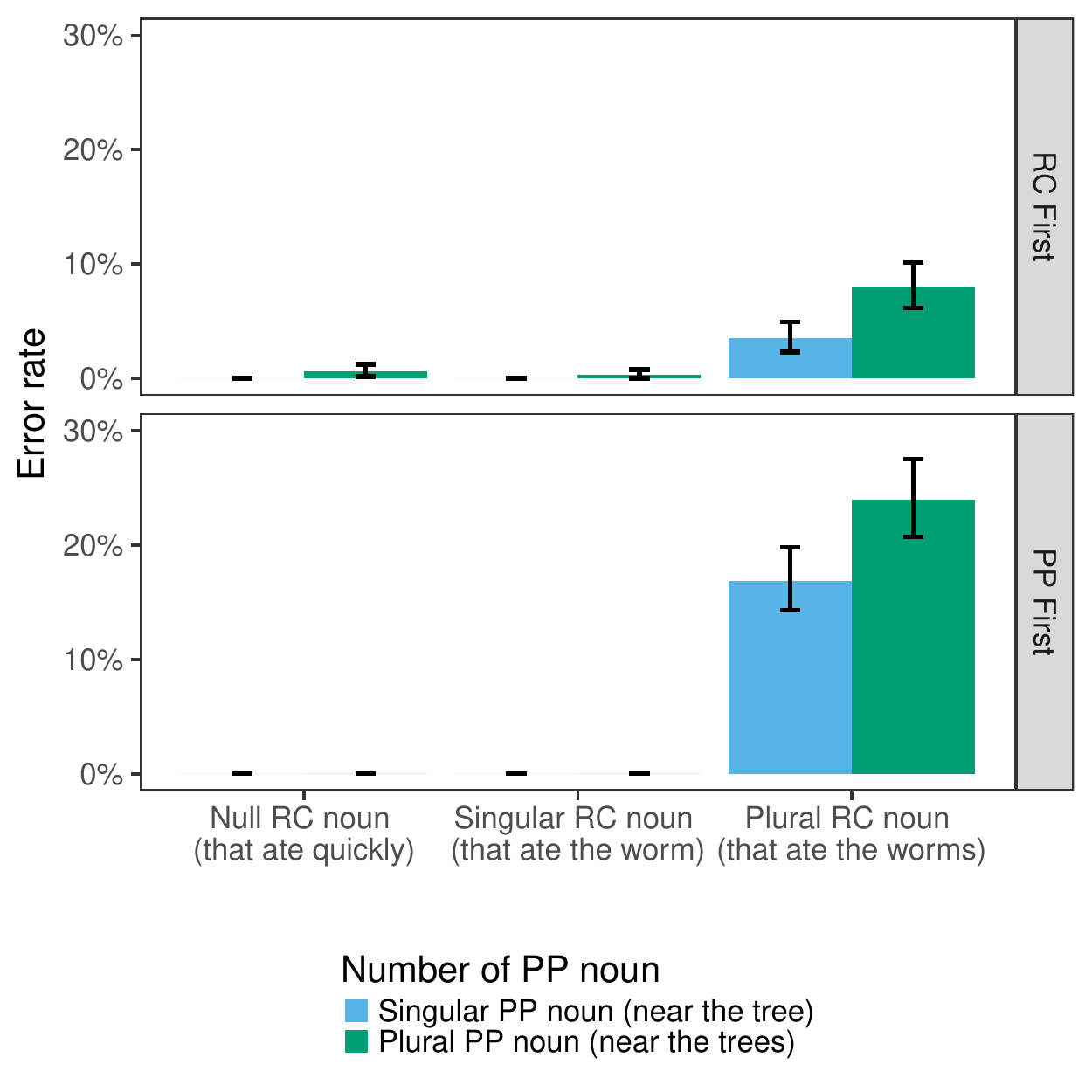}
        \label{fig:mult_flip_rnn}
    }
    \caption{(a) Human agreement errors in Experiment 2. Error bars represent bootstrapped 95\% confidence intervals. (b) RNN predictions for similar materials (top panel) and for the reversed version of the materials (bottom panel). Error bars represent standard errors (across the 20 models with a 1000-dimensional hidden layer).}
\end{figure*}
Each word of the preamble was encoded as a distributed representation (word embedding) and fed into an RNN with a single layer of long short-term memory (LSTM) units. The model did not have access to the characters that made up the word: the morphological relationship between \textit{table} and \textit{tables} was not indicated. The number of the verb was predicted from the final state of the RNN, after it has processed all of the words of the preamble. The distributed representations of the words and the weights of the recurrent layer and output layer were trained jointly.  We trained smaller models, in which the recurrent layer had 50 units (as in \citeNP{linzen2016assessing}), as well as larger models in which it had 1000 units. Word representations were 50-dimensional in both cases.\footnote{We trained 20 models of each size with different random initializations and report average performance. Optimization was performed using Adam. Training was stopped after an epoch in which the error on the validation set did not decrease (three epochs on average). We learned representations only for the 50000 most frequent words in the corpus (empirically, about 8 occurrences per million words or higher); less frequent words were replaced with their parts of speech (e.g., ``adverb''). }
\paragraph{Results:}

Five of the 32 items showed error rates exceeding 20\% across conditions, including the Match conditions. An examination of these items indicated that the unusual number of errors in these preambles was due to the presence of low frequency words.\footnote{When we sorted the items by the frequency of the subject or the attractor (the less frequent of the two), four of these five preambles were at the bottom of the list. The last outlier preamble (\textit{the rulers of the roman city - states}) did not have low frequency words. We conjecture that the unusually high error rate on this item was due to the presence of a hyphen, which may have been misinterpreted by the networks as introducing a parenthetical clause.} This suggests that the number representations that the network acquired for these words were not sufficiently robust; we excluded these five items from further consideration.

Figure \ref{fig:bc_rnn} shows the error patterns on the remaining items. The networks did not make any errors in preambles in which the subject and the local noun had the same number. This suggests that the RNNs learned to perform the task well and acquired appropriate representations of the number of the relevant words. Two aspects of the networks' error patterns are consistent with the human data: first, agreement errors were more common when the local noun did not match the subject in number; and second, these attraction errors were more likely when the subject was singular and the local noun plural than the other way around. Unlike humans, errors were much more likely when the attractor was embedded inside a relative clause; recall that the participants in Experiment 1 showed a moderate tendency in the opposite direction. 

Aside from this discrepancy, the general performance of the networks was very good: even in the condition in which performance was poorest, the error rate was 30\% (50-dimensional) or 24\% (1000-dimensional). The qualitative pattern of errors was identical between the 50-dimensional and 1000-dimensional models, with a slightly lower error rate for the 1000-dimensional models. This suggests that the 50-dimensional models evaluated by \citeA{linzen2016assessing} were large enough to capture the qualitative pattern.\footnote{In preliminary experiments, intermediate values for the size of the hidden layer (100, 250 and 500 units) did not lead to qualitatively different results.}

\section{Experiment 2}

We now turn to cumulative attraction. RNNs make more errors when there are two attractors than when there is one (Figure~\ref{fig:tacl}). Previous human studies have not directly compared sentences with one intervening noun to sentences with two. The most relevant previous work examined sentences with two intervening prepositional phrases and manipulated the number (singular or plural) of the nouns in those phrases, e.g., \textit{the threat to the president(s) of the compan(ies)} \cite{franck2002subject}. \citeauthor{franck2002subject} did not find a cumulative attraction effect. Perhaps surprisingly, the word closer to the subject (\textit{presidents}) was a more effective attractor than the word closer to the verb (\textit{companies}).

We adopted a different approach: instead of having two PPs as in \citeA{franck2002subject}, we used a relative clause with a PP embedded in it, e.g., \textit{The bird that ate the worm(s) near the tree(s)}. This made it possible to vary the number of intervening nouns, attractors or not, by replacing the first noun phrase (\textit{the worm}) with an adverb (\textit{quickly}). 

We constructed 36 critical items in six conditions and 76 filler items. Given that agreement attraction errors are much more likely with singular subjects, we limited ourselves to this configuration. We recruited 144 participants on Prolific, as before. We excluded 22 participants because their accuracy on the fillers was lower than 80\%. We used the self-paced reading paradigm of Experiment~1.

Figure~\ref{fig:mult_rep} shows the results of Experiment~2. Agreement errors were overall less frequent than in Experiment~1: even with two plural attractors, the average error rate was 12.7\%, compared to 22.6\% in the singular mismatch RC condition of Experiment~2. We do not have a definite explanation for this difference; one possible reason is that the words we used in Experiment~2 were more frequent than in Experiment~1.

The main determinant of error rates was the presence of a plural attractor at the beginning of the RC (\textit{that ate the worms}). There was no evidence of a cumulative attraction effect: \textit{the worms} was as strong an attractor as \textit{the worms near the trees}. In fact, the number of the second noun (\textit{tree/trees}) did not significantly affect error rates in any of the cases, in line with the finding that attraction in humans is not primarily caused by linear proximity to the verb \cite{franck2002subject}.

\section{Simulation of Experiment 2}

We extracted the predictions of our 1000-dimensional neural networks for materials based on Experiment~2 (see Footnote~\ref{modifiedmaterials} below). The results are shown in the top panel of Figure \ref{fig:mult_flip_rnn}. There were very few attraction errors when there was exactly one attractor and it was inside the PP (\textit{near the trees}), confirming that the RNNs were able to ignore nouns embedded in a PP. Errors increased when the noun directly following the beginning of the RC was an attractor (\textit{that ate the worms}). There was a cumulative attraction effect, with higher error rates when both nouns were plural.

Overall, the error rate was lower than in the corresponding conditions in Experiment~1. One possible explanation for this result is that the RC attractor in the current experiment is further from the verb than in Experiment~1. \citeA{linzen2016assessing} present evidence that RNNs use the plausible but flawed heuristic that RCs are usually short. If that is indeed the case, the RNNs may have concluded that the RC was over by the time they needed to make the prediction, and accordingly were better able to disregard the RC attractor.

To test this hypothesis, we constructed reversed versions of the materials (e.g., \textit{the bird near the trees that eats the worms}). To ensure that the sentences in the reversed condition were well-formed, we made certain minor changes to the materials of Experiment~2.\footnote{\label{modifiedmaterials}Typically, this involved changing the preposition (e.g., \textit{The captain that steered the ship through the storm} was changed into \textit{The captain that steered the ship in the storm}).} If the RNNs are relying on the short RC heuristic, we expect this reversal to increase the interference from the relative clause. Consistent with this expectation, the reversed sentences had more than double the error rate of the original ones. In contrast with humans, the attractor inside the PP was successfully ignored, even though it was closer to the subject. In other words, in RC-first sentences RNNs were similar to humans, but  for the wrong reason: those sentences confounded proximity to the subject (the reason for human errors) with the presence of an attractor inside an RC modifier (the reason for RNN errors).

\section{An examination of the short RC heuristic}

As another test of the hypothesis that our RNNs were relying on the heuristic that RCs tend to be short, we created a new set of materials:

\vspace{1 \baselineskip}

\begin{tabular}{@{}ll}
    \textit{Short:} & The lion that the tigers (ate) \\
    \textit{Medium:} & The lion that the hungry tigers (ate) \\
    \textit{Long:} & The lion that the extremely hungry tigers (ate)
\end{tabular}
\vspace{0.5 \baselineskip}

\begin{figure}
    \centering
    \includegraphics[width=0.8\textwidth]{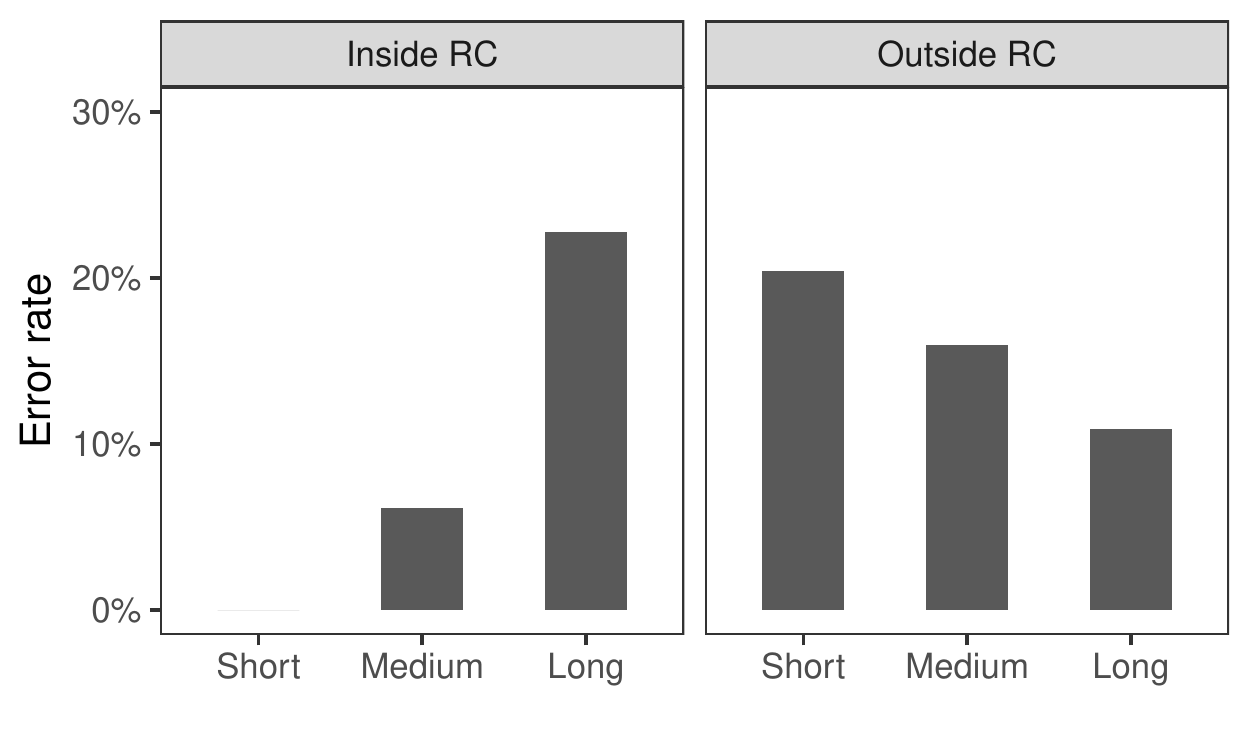}
    \caption{\label{fig:rc_length}Effect of relative clause length on error rate in predicting the number of the embedded verb (inside RC) and main verb (outside RC).}
\end{figure}

We probed the predictions of the network before and after \textit{ate}. In contrast with PPs, where the noun following the preposition can be safely ignored for the purpose of predicting agreement, sentences with RCs require attention to structure. Specifically, before \textit{ate} the verb should be plural, in agreement with the embedded subject \textit{tigers}. By contrast, since the verb \textit{ate} signals the end of the embedded clause, the verb that follows it should be singular, in agreement with the main clause subject \textit{lion}. This makes it impossible for the network to simply ignore the first noun after \textit{that}. Figure \ref{fig:rc_length} shows that the errors on these materials are consistent with the RC length heuristic. In the RC-external prediction point (after \textit{ate}), interference from the attractor decreases as the RC becomes longer. In the RC-internal condition (before \textit{ate}), the RNN increasingly reverts (incorrectly) to the main subject to make its prediction, even though the embedded subject \textit{tigers} is always immediately before the prediction point.

\section{Discussion}

Recurrent neural networks (RNNs) can be trained to predict with significant success whether a verb should be singular or plural, a challenging syntactic task thought to require structural representations. RNNs do make errors on this task, but so do humans. The experiments presented in this papers examined to what extent the detailed pattern of errors made by RNNs matches the errors made by humans.

While there were important similarities between RNNs and humans, which we discuss below, the error patterns differed in crucial ways. The RNNs were much more likely to make errors in sentences with relative clauses than in sentences without them, whereas the humans showed a small difference in the opposite direction (Experiment~1). Furthermore, errors in RNNs were affected by the proximity of the attractor to the verb rather than to the subject as in humans (Experiment~2). These findings suggest that the syntactic representations acquired by RNNs differ from those used by humans in sentence processing. In particular, RNNs are unable to detect the end of relative clauses in a categorical fashion; instead, they rely on a gradient heuristic whereby relative clauses are expected to be short, such that nouns that are further from the relativizer \textit{that} are less likely to be the embedded subject.

Both RNNs and humans made many more errors when the irrelevant noun had a different number from the subject than when it had the same number. This is unsurprising: if anything, an intervening noun with the same number as the subject should increase the probability of a correct verb number prediction. Another similarity between the RNNs and the human subjects is more noteworthy: in both cases, attraction errors were more likely when the subject was singular and the attractor plural than the other way around. Accounts of this asymmetry in humans have posited a semantic feature carried by plural nouns that can ``percolate'' up to the subject; singular nouns are assumed not to have such a feature, since singular is the \textit{unmarked} number.\footnote{The existence of attraction errors with plural subjects rules out a categorical version of this hypothesis, but a gradient version may still be viable.} The fact that this asymmetry was replicated in RNNs suggests that this feature system can emerge naturally in an RNN, perhaps because of the higher frequency of singular nouns in the language or due to frequency imbalances in noun number across syntactic constructions \cite{haskell2010experience}. In future work, these hypotheses can be tested by artificially manipulating the relative frequency of singular and plural nouns in various constructions in the training corpus.

There are at least two ways in which neural networks can be encouraged to make less errors on syntactically complex sentences. One way is by providing explicit syntactic annotations during training: the network is asked to predict not only the number of the verb but the entire structure of the sentence. In other work, we have shown that such a syntactic supervision signal can reduce agreement errors in sentences with relative clauses \cite{enguehard2017exploring}. Another way is to introduce stronger syntactic inductive biases into the architecture of the network. Crucially, RNNs made \textit{less} errors than humans in PPs; it remains to be seen if the modifications proposed in this paragraph would result in an \textit{increase} in agreement errors with PPs.

\bibliographystyle{apacite}

\setlength{\bibleftmargin}{.125in}
\setlength{\bibindent}{-\bibleftmargin}

\bibliography{rnn_rc_agr_cogsci}

\nocite{bock1992regulating}
\nocite{enguehard2017exploring}

\end{document}